# EngageTriBoost: Predictive Modeling of User Engagement in Digital Mental Health Intervention Using Explainable Machine Learning

Ha Na Cho
Department of Informatics
Donald Bren School of Information
University of California, Irvine
California, United States
Chohn1@uci.edu

Daniel Eisenberg
Department of Health Policy and Management, Fielding School of Public Health University of California, Los Angeles
Los Angeles, United States
Danieleisenberg@ucla.edu

Cheryl King
Department of Psychiatry and Psychology, Medical school
University of Michigan
Michigan, United States
Kingca@umich.edu

Kai Zheng
Department of Informatics,
Donald Bren School of Information
University of California, Irvine
California, United States
Kai.zheng@uci.edu

*Abstract*—Mental health challenges, particularly among young adults, are on the rise, necessitating effective solutions such as digital mental health interventions (DMHI). Despite their promise, DMHI face significant adoption barriers, including low initial uptake and high dropout rates. This study presents EngageTriBoost (ETB), an explainable ensemble machine learning (ML) framework designed to model multi-stage user engagement with *e*Bridge, a DMHI platform employing motivational interviewing-based online counseling. We trained and evaluated ETB on data from 1,673 at-risk college students, using 108 baseline features and engagement outcomes including initial login and subsequent message posting. ETB integrates XGBoost, LightGBM, and CatBoost as base learners with logistic regression as a meta-learner, tuned through cross-validation. Our approach emphasizes interpretability and transparency, rather than prioritizing raw predictive performance. ETB achieved up to 84% accuracy in predicting message posting, outperforming individual models in terms of recall and calibration. This framework demonstrated stable discrimination for message posting and more conservative performance for login prediction, showing inherent difficulty of predicting initial uptake. Using Shapley Additive exPlanations (SHAP), we characterized behavioral and demographic associations with engagement, including chronic pain, stigma, and alcohol use. Findings demonstrate the feasibility and value of explainable ML for optimizing DMHI engagement and guiding adaptive intervention strategies.

## I. Introduction

Mental health challenges, particularly depression and suicidal ideation, have significantly increased among young adults over the past few decades, creating an urgent public health concern [1]. Digital mental health intervention (DMHI) have emerged as a scalable, cost-effective tools to support mental health service utilization, reduce symptoms and increase access among underserved populations [2], [3]. Despite their promise, DMHI frequently suffer from high dropout [4], [5], and low engagement rates, with nonadherence reaching as high as 83%. Understanding and predicting user engagement remains a critical barrier to improving the efficacy of digital interventions [6], [7].

From 2015 to 2018, a multisite randomized controlled trial was conducted across four U.S. universities to evaluate the efficacy of the Electronic Bridge to Mental Health Services (*e*Bridge) [8], a DMHI designed to promote treatment uptake among college students at elevated risk for suicide. The intervention utilized personalized screening feedback and asynchronous counseling based on motivational interviewing, delivered through a secure web portal. While prior analyses showed that students who engaged with *e*Bridge were more likely to connect with mental health services, overall engagement remained low. Of the 1,673 students randomized to the intervention condition, only 627 (37.5%) logged into the *e*Bridge counseling platform at least once, 355 (21.0%) posted one message, and 168 (10.0%) posted two or more messages to the counselor [9]. These findings point to a missed opportunity for broader clinical impact and emphasize the importance of developing predictive models to identify factors associated with engagement.

In this study, we leverage data from the *e*Bridge trial to develop machine learning (ML) models that predict multi-level user engagement with the intervention. Prior research has applied ML to mental health problems [10], [11], but few have systematically addressed engagement prediction using diverse behavioral and demographic features, robust imputation strategies, and ensemble frameworks [12]. We introduce EngageTriBoost (ETB), a novel stacked ensemble design combining XGBoost, LightGBM, and CatBoost with logistic regression as a meta-learner. This approach improves both prediction and interpretability. Unlike prior work that treats engagement as a static binary outcome, we model multi-level behaviors (login, single message, multiple messages) to reflect varying degrees of involvement. This design supports finer-grained insights into user intent and commitment.

We further apply SHAP (Shapley Additive Explanations) for global and local interpretability. Compared to black-box feature importance metrics, SHAP enables consistent, theoretically grounded estimation of each feature's contribution to predictions. This is particularly valuable in mental health, where understanding individual variability and fairness is crucial for ethical deployment. Taken together, our contribution provides both a performance-driven and transparent framework for optimizing DMHI design and adaptive engagement interventions.

## II. RELATED WORK

### A. Machine Learning for DMHI Prediction

ML has been increasingly applied to DMHI for identifying symptom severity, predicting treatment outcomes, and detecting engagement patterns. Traditional models such as Logistic Regression or Random Forest have shown efficacy in early detection tasks [13], [14], [15], [16], [17]. More recent studies demonstrate strong performance of gradient boosting models such as XGBoost and LightGBM, achieving high accuracy in mental health classification tasks [18], [19], [20], [21]. However, these methods often rely on narrow feature sets and opaque architectures. Studies applying robust preprocessing and optimization techniques, including feature elimination, hyperparameter tuning, and cross-validation, further improve generalizability in health contexts [22], [23], however, few target engagement behavior in digital counseling settings.

### B. User Engagement Prediction using Machine Learning

Predicting engagement in DMHI is a rapidly emerging subfield, as researchers attempt to model behavioral adherence and platform use. Prior models based on static baseline variables, such as psychometric scores, financial hardship, or health burden, often fall short in capturing real-world engagement complexity [24]. Recent advances include just-in-time adaptive interventions [25], or probabilistic latent variable models to uncover five engagement profiles [26].

Demographic and structural factors also influence engagement. For instance, a study reported that race and ethnicity affected both income levels and subsequent willingness to engage with DMHI [27], while another study found that DMHI were particularly effective for older individuals, Asian users, and males [28]. In addition, explainable ML techniques have been applied to examine linguistic cues associated with engagement [29], though these efforts rarely quantified user perceptions of intervention credibility or value. Cross-platform dataset pooling multi-institutional data integration improves robustness in dropout prediction, with baseline depression shown to be a key barrier [30], [31]. Existing studies often predict engagement as a single binary outcome, ignoring behavioral nuance. Our study advances the field by using granular outcomes, a large multisite dataset, and interpretable ensemble methods to offer both predictive strength and actionable insights.

### C. Contribution of This Study

While prior work has demonstrated the utility of ML for digital mental health, much of the existing literature relies on limited feature sets, simplified engagement outcomes, and opaque model architectures. This study addresses these gaps by using a large randomized controlled trial of high-risk college students and modeling multi-stage engagement behaviors (login, single message, and sustained interaction) rather than treating it as a binary task. Second, we leveraged high-dimensional dataset incorporating over 100 behavioral, demographic, and psychological variables. We introduce ETB, a domain-tailored stacked ensemble combining complementary boosting algorithms, tailored to high-dimensional, self-reported survey data. Third, by comparing SHAP outputs at both global (cohort-wide) and local (individual-level) resolutions, we uncover behavioral, psychological, and social predictors of engagement. These interpretable outputs can inform both platform design and clinical outreach strategy. This approach improves over prior studies by balancing prediction accuracy, class imbalance mitigation, and transparent decision-making, providing a foundation for future real-time DMHI design and deployment.

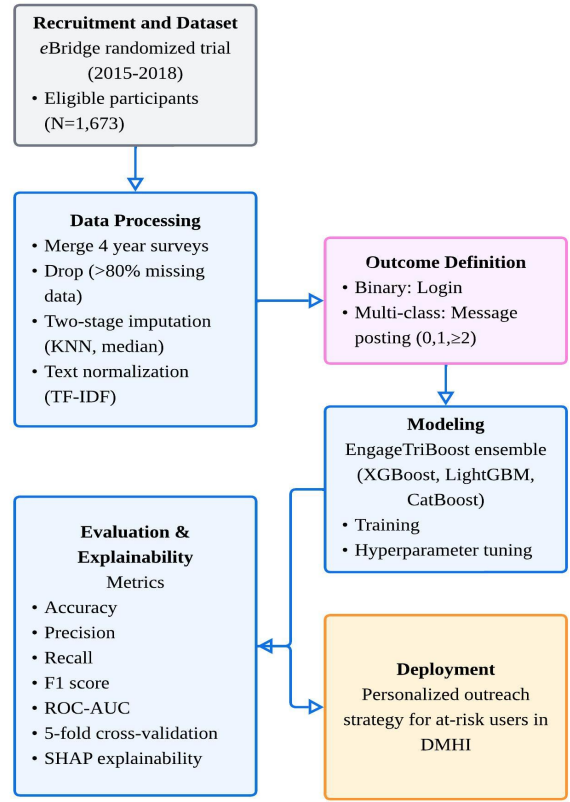

Fig. 1. Pipeline for predicting digital mental health engagement outcomes, starting from eBridge trial data preprocessing to modeling login and message posting outcomes using an ensemble approach with performance evaluation and SHAP explainability.

## III. METHODS

### A. Dataset and Outcomes

The empirical data collected from the *e*Bridge trial includes 108 predicting variables derived from baseline surveys administered to 1,673 college students randomized into the intervention. These variables span demographics, mental health indicators (e.g., depression, suicide ideation), substance use (e.g., alcohol, prescription misuse), perceived stigma, and behavioral risk factors. Based on the metrics proposed by Lipschitz et al. [32], engagement was assessed using three outcomes, (1) whether the participant logged into the counseling platform at least once (binary); (2) whether they posted one message (among those who logged in); and (3) whether they posted two or more messages. Outcomes (2) and (3) were modeled as multiclass ordinal classifications. Fig. 1 illustrate the full ML pipeline used in this study.

### B. Data Preprocessing

Survey data collected across four academic years were merged into a unified dataset with standardized variable names to ensure consistency. Features present across all years were retained, and duplicates were removed. To address missing data, features with more than 80% missingness were excluded to reduce noise from sparsely observed variables. Participant-level exclusion based on missingness was applied conservatively, where most participants had at least one missing value, and aggressive case deletion would substantially reduce representativeness. After filtering, the analytic dataset retained 108 predictor variables across all 1,673 participants. Missing values were addressed using two-stage imputation strategy: k-nearest neighbors imputation (k=5) followed by median imputation for any residual missingness. We set k=5 as a balance to reflect a trade-off between locality and stability that is commonly used in behavioral survey data.

Categorical variables were encoded using binary encoding for nominal features to reduce dimensionality while preserving information, and ordinal encoding for ordered variables to maintain hierarchical relationships. For example, biological sex was encoded as a binary variable (0=female, 1=male) based on self-reported sex at baseline. This strategy ensured efficient processing while retaining interpretability. Engagement outcomes were encoded numerically to standardize response variables. Login status was treated as a binary classification problem (1=logged in, 0=did not login). Message posting was structured as a multi-class classification problem (0=no message posted, 1=one message, 2=two or more messages), allowing models to distinguish varying engagement levels.

To standardize unstructured textual survey responses, we implemented a multi-step normalization process. First, Term Frequency-Inverse Document Frequency (TF-IDF) [33] vectorization (max 5,000 features; bi-gram range) converted textual responses into numerical feature vectors. These vectors were then grouped using a unsupervised clustering solely for dimensionality reduction and normalization. For text entries that did not fit well into any cluster, such as incomplete phrases, typos, or semantically ambiguous responses, a predefined rule-based mapping system was applied. This mapping referenced a curated dictionary of common variants and misspellings, assigning them to the nearest standardized category. This process reduced textual heterogeneity and improved reliability of downstream model input.

### C. Model Development

The dataset was split into an 80% training set and a 20% test set for model evaluation. Independent variables were standardized using feature scaling. To address class imbalance, the Synthetic Minority Over-sampling Technique (SMOTE) was applied to the training set, while test distributions were preserved. Oversampling was not stratified by protected attributes such as sex, race, or age. We acknowledge that non-stratified oversampling does not constitute a fairness-aware learning strategy and may amplify group patterns. Accordingly, model outputs are interpreted as associational signals.

Message-posting behavior was modeled as a multi-class classification problem with three levels: 0, 1, and two or more messages. Models used a softmax output layer with cross-entropy loss to estimate class probabilities. Performance metrics included macro-averaged precision, recall, and F1-score to ensure equal weighting across classes despite imbalance.

Although login precedes message posting, we modeled the two outcomes independently to identify distinct predictors for initial uptake versus sustained engagement. Future work will incorporate hierarchical or sequential approaches when longitudinal timestamps become available.

Seven baseline classifiers were implemented: Logistic Regression (LR), Decision Tree (DT), Random Forest (RF), Support Vector Machine (SVM), Extreme Gradient Boosting (XGBoost) [34], Light Gradient Boosting Machine (LightGBM) [35], and Categorical Boosting (CatBoost) [36]. Feature selection used univariate scoring, with hyperparameters tuned via randomized grid search and five-fold cross-validation.

### D. Ensemble Learning: EngageTriBoost

To improve robustness and capture heterogeneous engagement behaviors, we developed ETB, a domain-specific stacked ensemble model. The ETB architecture integrates XGBoost, LightGBM, and CatBoost as base learners with logistic regression classifier serving as the meta-learner. Each base learner processes the same training data and independently generates class probabilities. These out-of-fold predictions are then used as input features to train the meta-learner, which learns optimal combinations of base model outputs to make the final prediction. This stacked architecture combines complementary strengths: XGBoost's strong regularization and generalization in sparse, noisy data, LightGBM's speed and scalability for large feature spaces through leaf-wise tree growth, and CatBoost's native handling of categorical variables and ordered boosting, reducing prediction shift. The meta-learner (LR) was chosen for its transparency and low variance, enabling interpretability of how each base model contributes to the final prediction. To prevent overfitting and data leakage, we implemented stacked generalization using five-fold cross-validation. Base learners were trained within each fold, and their predictions were used to train the meta-learner on held-out data only. Hyperparameter tuning for each base model was conducted independently on training folds using randomized search, and final configurations are reported in the Results section. This architecture is particularly suitable for high-dimensional, self-reported survey datasets common in DMHI research, where single models often overfit or underperform due to feature noise and class imbalance. By combining multiple learners and adding a second-layer model for consensus, ETB enhances both predictive stability and individual-level interpretability, which are crucial for real-world application in digital mental health.

TABLE I. PERFORMANCE COMPARISON FOR MACHINE LEARNING MODELS ON BINARY CLASSIFICATION OF USER LOGIN BEHAVIOR.

| Model | Accuracy (CI) | ROC-AUC (CI) | Precision (CI) | Recall (CI) | F1 (CI) |
|---|---|---|---|---|---|
| LR | 0.60 (±0.02) | 0.60 (±0.03) | 0.41 (±0.03) | 0.60 (±0.04) | 0.48 (±0.03) |
| DT | 0.62 (±0.03) | 0.57 (±0.04) | 0.44 (±0.04) | 0.34 (±0.05) | 0.39 (±0.04) |
| RF | 0.65 (±0.02) | 0.66 (±0.03) | 0.53 (±0.03) | 0.28 (±0.04) | 0.37 (±0.03) |
| SVM | 0.61 (±0.03) | 0.59 (±0.04) | 0.45 (±0.03) | 0.65 (±0.05) | 0.54 (±0.04) |
| XGB | 0.67 (±0.02) | 0.65 (±0.03) | 0.52 (±0.03) | 0.18 (±0.03) | 0.27 (±0.02) |
| LightGBM | 0.67 (±0.02) | 0.66 (±0.03) | 0.55 (±0.03) | 0.30 (±0.04) | 0.38 (±0.03) |
| CatBoost | 0.65 (±0.02) | 0.66 (±0.03) | 0.44 (±0.03) | 0.21 (±0.03) | 0.28 (±0.02) |
| ETB | 0.64 (±0.02) | 0.67 (±0.03) | 0.50 (±0.03) | 0.40 (±0.04) | 0.44 (±0.03) |

TABLE II. PERFORMANCE COMPARISON ON MULTI-CLASS CLASSIFICATION RESULTS OF MESSAGE POSTING BEHAVIOR. FOR MULTICLASS OUTCOMES, PRECISION, RECALL, AND F1 SCORES ARE MACRO-AVERAGED.

| Model | Accuracy (CI) | ROC-AUC (CI) | Precision (CI) | Recall (CI) | F1 (CI) |
|---|---|---|---|---|---|
| LR | 0.42 (±0.03) | 0.56 (±0.04) | 0.77 (±0.03) | 0.79 (±0.03) | 0.78 (±0.02) |
| DT | 0.56 (±0.03) | 0.50 (±0.05) | 0.75 (±0.03) | 0.53 (±0.04) | 0.62 (±0.03) |
| RF | 0.79 (±0.02) | 0.54 (±0.03) | 0.76 (±0.03) | 0.80 (±0.03) | 0.78 (±0.02) |
| SVM | 0.76 (±0.02) | 0.50 (±0.04) | 0.75 (±0.03) | 1.00 (±0.01) | 0.86 (±0.02) |
| XGB | 0.80 (±0.02) | 0.54 (±0.03) | 0.75 (±0.03) | 1.00 (±0.01) | 0.86 (±0.02) |
| LightGBM | 0.80 (±0.02) | 0.54 (±0.03) | 0.75 (±0.03) | 0.93 (±0.02) | 0.93 (±0.01) |
| CatBoost | 0.67 (±0.02) | 0.56 (±0.04) | 0.76 (±0.03) | 0.93 (±0.02) | 0.84 (±0.02) |
| ETB | 0.84 (±0.01) | 0.52 (±0.03) | 0.82 (±0.02) | 1.00 (±0.00) | 0.90 (±0.01) |

### E. Evaluation and Explainability

Model performance on the held-out test set was assessed using accuracy, macro-averaged precision, recall, F1-score, and ROC-AUC. Macro-averaged metrics were applied to the multi-class message-posting task (three levels: 0, 1, and two or more messages), while binary outcomes such as login prediction used the same metrics directly. To account for variability in test set estimates, 95% confidence intervals (CI) were computed via 1,000 bootstrapped resamples for each metric. ETB was benchmarked against all baseline models under identical evaluation criteria.

Model interpretability was provided by SHAP (Shapley Additive exPlanations) [37], which quantified feature contributions at both global and local levels. For message-posting predictions, an additional binary comparison (0 vs. ≥1 messages) highlighted predictors linked to the transition from non-engagement to active engagement. SHAP summary and decision plots identified behavioral, demographic, and psychological factors influencing engagement patterns.

## IV. RESULTS

### A. Participant Description

A total of 1,673 students were randomized to the intervention of the *e*Bridge study and included in this analysis. The sample primarily comprised undergraduate students, with a majority identifying as female (61.6%) and the remainder identifying as male (38.4%) or small proportion identifying with other sex identities. Participants were predominantly aged 18 to 22, with approximately 20% aged 23 and older. Racially, the sample was largely White (72.0%), followed by Asian (21.0%), Black (7.4%), and students from other racial and ethnic backgrounds (3.6%). Engagement with the *e*Bridge platform was modest overall. Post-randomization, 37.5% of students logged in at least once. Among those who logged in, 21.2 % posted one message to a counselor, while 10.0% posted two or more messages. These behavioral outcomes served as the target variables for the machine learning model development, and were treated as separate classification tasks.

### B. Model Performance

Predicting initial login proved more challenging than predicting message posting across all models. Login reflects a threshold decision influenced by unobserved contextual factors such as motivation, stigma, and competing demands. Table I presents the classification performance of eight machine learning models on predicting user login behavior. Among all models, ETB achieved the highest ROC-AUC (0.67, 95% CI: 0.64-0.70), and a balanced F1 score (0.44, 95% CI: 0.41-0.47) through improved trade-offs in precision (0.50, 95% CI: 0.47-0.53) and recall (0.40, 95% CI: 0.37-0.43). While XGBoost,

LightGBM, and CatBoost, matched ETB in accuracy (0.67), they exhibited lower recall and F1 scores, suggesting reduced sensitivity to correctly identifying engaged users. In contrast, ETB maintained a better balance between precision (0.50) and recall (0.40), resulting in a respectable F1 score of 0.44. In comparison, RF and LightGBM achieved higher precision (0.53 and 0.55, respectively), while LR and SVM yielded higher recall (0.60 and 0.65, respectively). However, these models showed lower F1 scores due to trade-offs in either sensitivity or specificity. ETB thus provided a moderately balanced predictive performance, without extreme skew in either direction, which is beneficial for real-world engagement detection where both false positives and false negatives carry risk.

Table II summarizes the model performance for predicting message posting behavior, categorized into three levels: 0, 1, or ≥2 messages. ETB outperformed all baselines in accuracy (0.84, 95% CI: 0.81-0.87), precision (0.82, 95% CI: 0.79-0.85), recall (1.00, 95% CI: 0.97-1.00), and F1 score (0.90, 95% CI: 0.88-0.93). Although several models such as SVM, XGBoost, and LightGBM achieved perfect recall, their F1 scores were lower due to reduced precision. Notably, ETB maintained high recall without compromising precision, demonstrating robustness in identifying users with higher engagement levels. This performance highlights ETB's effectiveness in handling class imbalance while preserving sensitivity and specificity across multi-class outcomes. All reported metrics in Table I and Table II are from the held-out test set only.

All models underwent hyperparameter tuning grid search with five-fold cross-validation to maximize predictive performance. For ETB, which integrates XGBoost, LightGBM, and CatBoost the optimal configuration for message posting prediction included learning rate of 0.01, maximum depth of 9, and 200 estimators for each base learner, reflecting a low-bias, high-capacity ensemble. For the login prediction task, individual boosting models were tuned more granularly. XGBoost achieved optimal results with a learning rate of 0.05, a maximum depth of 9, 300 estimators, a column sampling rate by tree of 0.8, a subsample rate of 0.8, and no regularization via gamma. LightGBM was configured with a learning rate of 0.05, a maximum depth of 6, 200 estimators, 50 leaves, and the same sampling parameters as XGBoost. CatBoost achieved optimal performance with a learning rate of 0.05, a depth of 6, 200 iterations, an L2 leaf regularization value of 3, and verbosity disabled to reduce logging output during training.

Baseline models were also tuned for fair comparison. The LR model employed L1 regularization, a regularization strength (C) of 0.1, and the liblinear solver. The DT classifier achieved optimal performance with a maximum depth of 10, a minimum of 5 samples per leaf, a minimum of 2 samples per split, the entropy criterion, and a random splitter. The RF model was configured with a maximum depth of 20, 100 estimators, and default splitting settings. Lastly, the SVM was tuned using regularization parameter (C) of 10. These hyperparameter optimizations contributed to each model's generalization ability and overall robustness in predicting digital engagement outcomes.

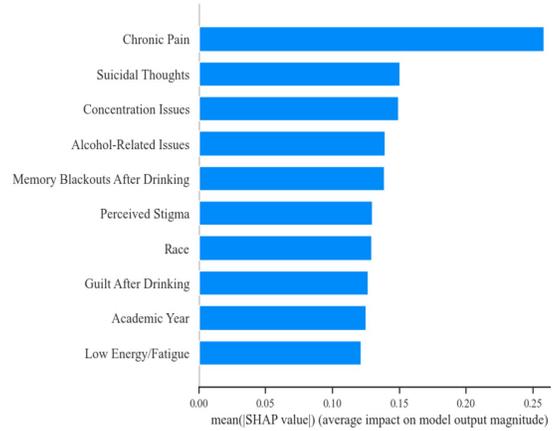

(a) Mean SHAP value plot

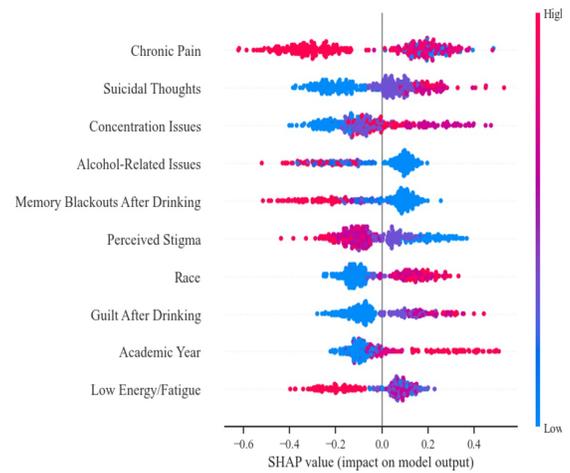

(b) SHAP summary plot

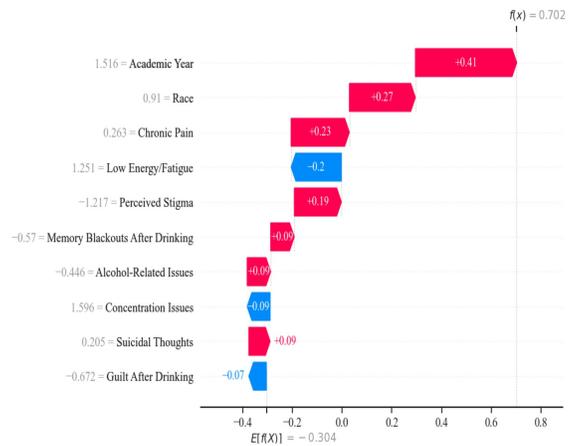

(c) SHAP decision plot

Fig. 2. SHAP explanations of the top ten features influencing user login behavior. (a) Mean absolute SHAP values, representing the global feature imporatnce. (b) Summary plot feature value (red=high, blue=low) and SHAP impact (right=increased, left=decreased login). (c) Decision plot visualizing the cumulative contribution of key features in representative individual prediction.

## C. Model Interpretation

The SHAP values quantify feature contributions within the trained model and reflect associations rather than causal effects; all interpretations are therefore framed in terms of increased or decreased model attribution rather than direct behavioral causation. The SHAP feature importance plots in Fig. 2 illustrate key contributors to login prediction. The global summary plot in Fig. 2.a. shows that higher chronic pain values were associated with increased SHAP attribution toward login predictions, whereas lower pain values were associated with decreased attribution toward login. This pattern is visually reinforced by the rightward shift and red coloration in the SHAP summary bars for chronic pain, indicating positive SHAP values pushing the model toward a login output.

Elevated suicidal ideation and difficulty concentrating also exhibited positive SHAP attribution toward login predictions. For instance, the concentration difficulty feature shows a high SHAP magnitude with a predominantly red color gradient, further supporting its positive contribution to the model output. In contrast, perceived stigma was associated with negative SHAP attribution toward login prediction. Higher stigma values corresponded to blue SHAP contributions in the global plot, indicating reduced attribution toward login, with broader tails suggesting heterogeneity across individuals.

Alcohol-related features demonstrated bidirectional effects. Frequent blackouts and higher overall alcohol consumption were associated with negative SHAP attribution toward login predictions (Fig. 2.a.), whereas feelings of guilt or remorse after drinking showed positive SHAP attribution. Demographic features also exhibited differential SHAP attribution. White race users was associated with higher SHAP attribution toward login predictions compared with Black, American Indian, and Pacific Islander race users, indicating potential disparities in modeled engagement signals. Undergraduate students showed higher attribution toward login relative to graduate students, while low self-reported energy or fatigue was associated with negative SHAP attribution.

Local SHAP decision plots in Fig. 2.c. provide individual-level explanations, illustrating how specific feature values contributed toward login or non-login predictions for representative cases. For example, one individual with high chronic pain and suicidal ideation showed strong positive SHAP contributions toward login, whereas another individual with elevated stigma and low distress symptoms showed negative contributions. While global SHAP values summarize average marginal effects, local plots reveal case-specific patterns that may diverge, highlighting heterogeneity in engagement signals across individuals.

In Fig. 3, SHAP explanations reveal the primary drivers of message posting predictions. Sex was the strongest attribution, with female users associated with increased SHAP attribution toward message posting. Several alcohol-related features, including higher drinking frequency, difficulty controlling alcohol use, and prior advisement to reduce consumption, demonstrated positive SHAP attribution toward message posting, suggesting awareness of substance-related concerns contribute to modeled engagement signals and may actively seek peer or counselor support via messages.

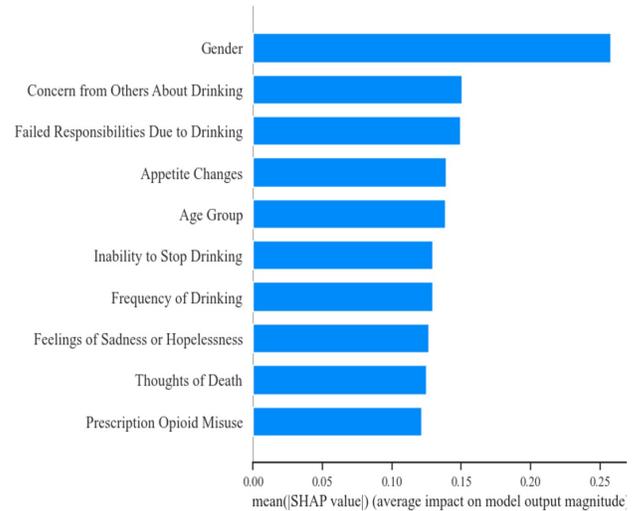

(a) Mean SHAP value plot

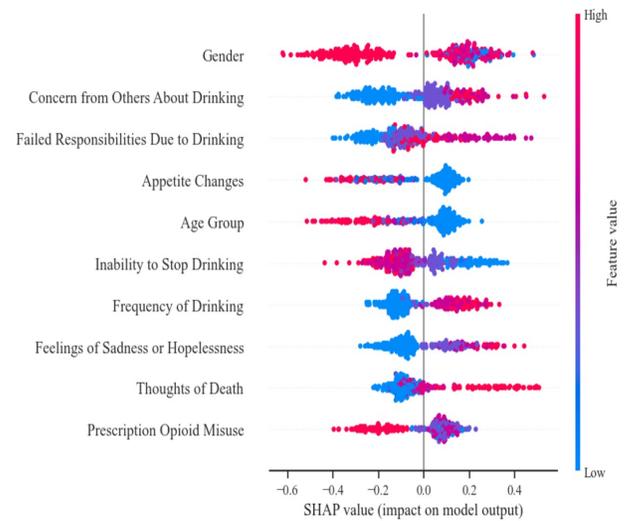

(b) SHAP summary plot

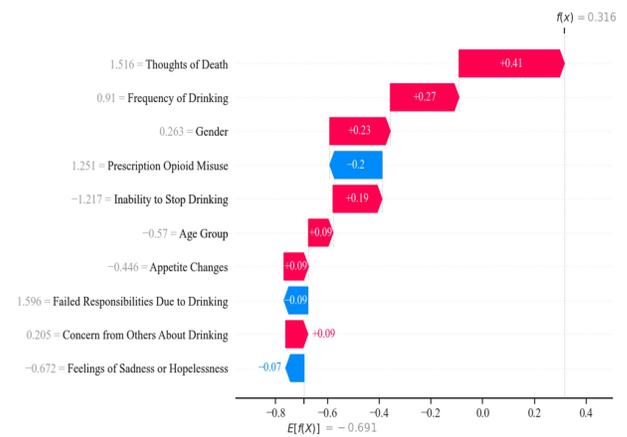

(c) SHAP decision plot

Fig. 3. SHAP explanations for top ten features influencing message posting behavior, presented using the same layout and interpretation as in Fig. 2.

Appetite fluctuations, both reduced and increased, were associated with elevated SHAP attribution toward message posting activity, implying emotional or physiological dysregulation. Older age was associated with increased SHAP attribution toward message posting. Persistent sadness, hopelessness, and prolonged suicidal ideation also showed positive SHAP attribution toward message posting predictions, whereas nonmedical use of prescription pain medications exhibited negative attribution.

Across both figures, feature directionality (red indicating increased attribution, and blue indicating decreased attribution) and SHAP magnitude provide interpretable insights into how behavioral, psychological, and demographic factors contribute to modeled engagement pathways- initial access (login) versus sustained interaction (message posting). The combination of interpretable ensemble modeling and SHAP visualization supports transparent analysis of engagement signals in DMHI design.

## V. DISCUSSION

Our ETB's ensemble design offers a structured integration of three boosting models, harnessing their individual strengths while mitigating their weaknesses. Unlike single-model classifiers, ETB consistently achieved superior F1 scores across both binary and multi-class tasks, maintaining equilibrium between recall and precision. This stability is crucial in digital mental health platforms, where misclassification can lead to missed intervention opportunities or excessive false alerts.

Notably, ETB achieved its strongest performance on the multi-class prediction of message posting. This outcome is expected given the behavioral distinction between login (initial access) and message posting (sustained interaction) [38]. Login behavior is more volatile, shaped by contextual barriers like time constraints or tech access, resulting in noisier signals. In contrast, message posting captures sustained, voluntary engagement, which is more closely tied to stable psychological and behavioral indicators. These structural differences in the underlying data distribution likely contributed to the clearer class boundaries learned by ETB, enabling more reliable differentiation between levels of engagement.

ETB's balanced performance arises from leveraging complementary strengths of its integral models, resulting in enhanced robustness across diverse engagement behaviors. Unlike standalone classifiers, it maintained more consistent recall and precision, indicating a reduced risk of false negatives and false positives. Importantly, it demonstrated improved generalizability through integration of data from multiple institutions, showing performance comparable to prior applied models under similar real-world constraints [39], [40].

SHAP analysis revealed interpretable, individualized insights, identifying emotional, demographic, and behavioral factors that influence engagement. Emotional distress indicators (e.g., suicidal ideation, trouble concentrating) positively influenced login, highlighting passive support-seeking, whereas persistent sadness and long-term ideation predicted message posting-indicative of deeper, communicative coping needs. These results align with prior findings on user intent and emotional state in DMHI engagement.

Demographic attributes significantly moderated engagement in distinct ways. Undergraduate students consistently exhibited higher login rates than graduate students, which may reflect differences in help-seeking behavior and perceived need for support. Older users showed greater message activity, potentially demonstrating increased self-awareness or comfort with emotional disclosure. Sex differences aligned with established patterns, as female users posted more messages, while male users exhibited less expressive engagement, reinforcing the need for sex-sensitive design in digital interventions. Racial disparities in engagement were also evident, with White users logging in more frequently than underrepresented groups (e.g., Black, Native American, and Pacific Islander users). These disparities highlight the urgency of culturally inclusive approaches to platform design.

Behavioral health factors, particularly those related to alcohol use and chronic pain, revealed complex engagement patterns. Rather than uniformly deterring or promoting platform use, alcohol-related features demonstrated emotionally differentiated effects. High alcohol consumption and memory blackout episodes were associated with reduced logins, possibly showing avoidance or lower self-efficacy during periods of risky behavior. In contrast, feelings of guilt and difficulty controlling drinking were strongly tied to message posting, inferring that self-reflective states or external feedback may catalyze communicative engagement. These findings point to an opportunity where internal conflict or behavioral insight drives users to seek DMHI support. Chronic pain, while associated with increased login frequency, did not predict message posting. This pattern may reflect monitoring behavior among users managing ongoing physical discomfort, stressing a potential need to design features that support expressive interaction within this subgroup.

These trends point to a critical dichotomy between passive and active engagement. While login behaviors reflect acute emotional and behavioral cues, message posting corresponds to sustained distress and proactive coping. Recognizing this distinction supports passive users through timely prompts and providing interactive channels for those experiencing deeper or prolonged challenges. ETB can serve as an early adaptive monitoring module, flagging users at risk of disengagement in near real-time to support just-in-time interventions.

While the findings provides promising evidence, several limitations should be acknowledged. First, our model has not yet undergone external validation. Future validation on independent digital mental health datasets will be critical to evaluate the model's generalizability and ensure consistent performance across diverse populations, as well as to determine whether model recalibration or feature adaptation is required for broader deployment. Second, engagement was modeled as a static outcome. Future work should incorporate sequence modeling to capture trajectory-based patterns. Third, SHAP explains associations rather than causal relationships and may be affected by feature correlations [41]. Advancing this framework entails applying temporal models to track engagement evolution (e.g., LSTM, Bayesian networks), leveraging real-time behavioral signals through digital phenotyping (e.g., smartphone behavior or sensor data), and integrating hybrid frameworks that blend AI predictions with human-in-the-loop interventions. These additions can improve accuracy, responsiveness, and equity in identifying and addressing disengagement risk.

From an ethical perspective, our work underscores the need for equity-aware ML pipelines. Bias mitigation must begin during data preprocessing (e.g., balancing racial and gender representation) and continue through performance stratification. Privacy safeguards (e.g., de-identification, HIPAA compliance) and informed consent protocols are essential, especially when handling sensitive psychological data [42], [43]. As large-scale, AI-enabled interventions evolve, ongoing attention will be needed to fairness audits, model interpretability tools, and culturally adaptive design principles must be incorporated to ensure transparency and inclusivity. We also emphasize the importance of stakeholder literacy. Clinicians, patients, and developers must understand model limits to avoid overreliance and promote safe, human-centered deployment.

## VI. Conclusion

This study presents a machine learning framework for predicting multi-level engagement in DMHI. The proposed ensemble model, ETB, demonstrated strong predictive performance and, through SHAP analysis, identified key behavioral and demographic factors influencing engagement. These findings provide actionable insights for designing equitable and user-centered digital mental health strategies. When integrated into real-world DMHI platforms, ETB could function as a real-time risk flagging system, enabling timely and personalized interventions, such as automated reminders or counselor outreach, for individuals at risk of disengagement, with future work focusing on external validation and longitudinal modeling to enhance generalizability and clinical impact.